\pdfoutput=1
\documentclass[10pt,twocolumn,letterpaper]{article}

\usepackage{wacv}
\usepackage{times}
\usepackage{epsfig}
\usepackage{graphicx}
\usepackage{amsmath}
\usepackage{amssymb}
\usepackage{booktabs}
\usepackage{authblk}
\usepackage[accsupp]{axessibility}  % Improves PDF readability for those with disabilities.

\usepackage{multirow}

\usepackage{cite}

\def\wacvPaperID{15} % WACV workshop paper ID

\wacvalgorithmstrack   % Uncomment this line if you are submitting to the Algorithms Track.
\wacvfinalcopy % *** Uncomment this line for the final submission

\ifwacvfinal
\usepackage[breaklinks=true,bookmarks=false]{hyperref}
\else
\usepackage[pagebackref=true,breaklinks=true,colorlinks,bookmarks=false]{hyperref}
\fi

\begin{document}

%%%%%%%%% TITLE
%\title{Transformer-based Multiple Granularity Features \\ for Unsupervised Person Re-Identification}

\title{Transformer Based Multi-Grained Features \\ for Unsupervised Person Re-Identification}

% \author{Jiachen Li\\
% \\
% % Zhejiang University\\
% {\tt\small lijiachen\_isee@zju.edu.cn}\\
% % {\tt\small firstauthor@i1.org}
% % For a paper whose authors are all at the same institution,
% % omit the following lines up until the closing ``}''.
% % Additional authors and addresses can be added with ``\and'',
% % just like the second author.
% % To save space, use either the email address or home page, not both
% \and
% Menglin Wang\\
% Zhejiang University\\
% {\tt\small lynnwang6875@gmail.com}
% \and
% Xiaojin Gong\\
% \\
% % Zhejiang University\\
% % Institution2\\
% % First line of institution2 address\\
% {\tt\small gongxj@zju.edu.cn}
% }

\author[1]{Jiachen Li}
\author[2]{Menglin Wang}
\author[ ]{Xiaojin Gong\thanks{The corresponding author.}}
\affil[1,2,$\ast$]{College of Information Science and Electronic Engineering, Zhejiang University, China}
\affil[ ]{{\tt\small $^{1,*}$\{lijiachen\_isee, gongxj\}@zju.edu.cn, $^2$lynnwang6875@gmail.com;}}

\maketitle
\thispagestyle{empty}

%%%%%%%%% ABSTRACT
\begin{abstract}
Multi-grained features extracted from convolutional neural networks (CNNs) have demonstrated their strong discrimination ability in supervised person re-identification (Re-ID) tasks. Inspired by them, this work investigates the way of extracting multi-grained features from a pure transformer network to address the unsupervised Re-ID problem that is label-free but much more challenging. To this end, we build a dual-branch network architecture based upon a modified Vision Transformer (ViT). The local tokens output in each branch are reshaped and then uniformly partitioned into multiple stripes to generate part-level features, while the global tokens of two branches are averaged to produce a global feature. Further, based upon offline-online associated camera-aware proxies (O2CAP) that is a top-performing unsupervised Re-ID method, we define offline and online contrastive learning losses with respect to both global and part-level features to conduct unsupervised learning. Extensive experiments on three person Re-ID datasets show that the proposed method outperforms state-of-the-art unsupervised methods by a considerable margin, greatly mitigating the gap to supervised counterparts. Code will be available soon at \url{https://github.com/RikoLi/WACV23-workshop-TMGF}.
\end{abstract}

%%%%%%%%% BODY TEXT
\section{Introduction}
Purely unsupervised person re-identification (Re-ID) aims to learn a Re-ID model without using any identity labels. This task has attracted extensive research interest because of its label-free manner, which makes it more practical and scalable to real-world deployments. Over the past few years, significant progress has been made, mainly due to the leverage of pseudo labeling~\cite{BUC} and contrastive learning~\cite{wu2018unsupervised,SimCLR,MoCo} techniques. Existing unsupervised methods often focus on the design of various contrastive losses~\cite{CAP,dai2021cluster,ICE,O2CAP} and the refinement of noisy pseudo labels~\cite{O2CAP,MGH,RLCC,PPLR}. Most of them pay little attention to the improvement of their feature extraction networks, which are crucial for identification as well. 

On the contrary, the architecture of feature extraction backbones has been extensively investigated in supervised person Re-ID. For example, besides bag of tricks (BoT)~\cite{BoT}, partition-based~\cite{PCB,Cheng2016} or multi-granularity~\cite{MGN,zheng2019pyramidal} networks are developed to capture fine-grained cues, and attention schemes~\cite{Li2018HAN,Si2018Dual,ABD-Net} are integrated to concentrate on discriminative parts. Recently, self-attention or transformer mechanisms~\cite{APD,Zhang2021HAT,PAT,AAformer,LA-Transformer,TransReID,TransReID-SSL} have also been successfully applied to supervised Re-ID. Some of them~\cite{APD,Zhang2021HAT,PAT} integrate transformers with convolutional neural networks (CNNs) and the others~\cite{AAformer,LA-Transformer,TransReID,TransReID-SSL} construct pure transformer architectures to explore long-range contexts. It has been validated that both fine-grained cues and long-range contexts greatly boost the performance of supervised Re-ID.

Inspired by the techniques developed in supervised Re-ID, especially by the CNN-based Multiple Granularity Network (MGN)~\cite{MGN} and the pure transformer networks~\cite{TransReID,TransReID-SSL,AAformer,LA-Transformer}, we intend to investigate the way of extracting multi-grained features from a pure transformer network to address the more challenging unsupervised Re-ID problem. To this end, we take the TransReID-SSL~\cite{TransReID,TransReID-SSL} network as our backbone, which builds upon the Vision Transformer (ViT)~\cite{ViT} but slightly modifies ViT to adapt to the Re-ID task. Then, we construct a dual-branch architecture based on the transformer backbone. Each branch duplicates the last transformer layer to produce outputs independently. The output local tokens in each branch, which are corresponding to input patches~\cite{LA-Transformer,Beal2020}, are reshaped and uniformly partitioned into a number of stripes to learn part-level features. Meanwhile, the output global tokens of both branches are averaged to generate a global feature. By this means, multi-grained features are effectively learned from the pure transformer network.

In order to take advantage of the learned multi-grained features for unsupervised Re-ID, we choose offline-online associated camera-aware proxies (O2CAP)~\cite{O2CAP} as our learning framework. O2CAP~\cite{O2CAP} is a state-of-the-art unsupervised Re-ID method based on a CNN backbone. We extend it by replacing its backbone with our transformer-based dual-branch network for feature extraction. In addition, we define offline and online contrastive learning losses with respect to both global and part-level features. 

%Moreover, besides the original losses defined in O2CAP, we additionally define offline and online contrastive learning losses with respect to part-level features.

%In addition, we define offline and online contrastive learning losses with respect to both global and part-level features. 

Although some recent works also attempt to learn part-level features from pure transformer networks~\cite{AAformer,LA-Transformer} for supervised Re-ID or leverage CNN-based part-level features for unsupervised Re-ID~\cite{Yang2019_patch,SSG,UP-ReID,Lin2020_part,PPLR}, our method distinguishes itself from them in the following aspects:
\begin{itemize}
	\item Inspired by MGN~\cite{MGN}, we design a dual-branch architecture appended to a pure transformer network to learn features at multiple granularities, which is simple but effective to mine fine-grained cues and capture long-range contexts at the same time.  
	\item Based upon O2CAP~\cite{O2CAP}, we additionally define part-feature based offline and online contrastive learning losses to leverage the learned multi-grained features for unsupervised Re-ID, boosting the performance significantly. 
	\item Extensive experiments on three person Re-ID datasets show that our method outperforms state-of-the-art unsupervised methods by a considerable margin, greatly mitigating the gap to supervised counterparts. 
\end{itemize}

\section{Related Work}
\subsection{Unsupervised Person Re-ID}
Previous unsupervised Re-ID methods can be roughly grouped into unsupervised domain adaptation (UDA)-based~\cite{SpCL,MCRN,GLT,feng2021complementary,DARC} or purely unsupervised~\cite{BUC,HCT,ICE,IICS,CAP,dai2021cluster,O2CAP} categories. In recent years, the purely unsupervised Re-ID has attracted more research interest due to its promising performance and no use of extra labeled datasets. Most unsupervised methods are clustering-based, taking clustering to produce pseudo labels and training a Re-ID model under the supervision of pseudo labels iteratively. Advanced performance has been recently achieved via the design of clustering techniques~\cite{HCT,zheng2021online}, the leverage of contrastive learning techniques~\cite{CAP,dai2021cluster,O2CAP}, the refinement of noisy pseudo labels~\cite{MGH,RLCC,PPLR}, etc. Although numerous methods have been developed, almost all of them leave their CNN-based feature extraction backbones unchanged. An exception is TransReID-SSL~\cite{TransReID-SSL}, which attempts to exploit a pure transformer network for unsupervised Re-ID and significantly boosts the performance.

%{\color{blue}An exception is TransReID-SSL~\cite{TransReID-SSL}, which attempts to exploit transformer-based backbone to address the unsupervised Re-ID task and significantly boosts the performance.} 

%For instance, CAP~\cite{CAP}, ICE~\cite{ICE}, MGH~\cite{MGH}, CC~\cite{dai2021cluster} and O2CAP~\cite{O2CAP} conduct contrast at different clustering levels and design various contrastive losses for learning. Most methods focus on the design of contrastive losses and the refine of pseudo labels, while leaving the CNN-based feature extraction backbones unchanged. except TransReID-SSL~\cite{TransReID-SSL}, 

\subsection{Part-based Person Re-ID}
Part-based features can encode fine-grained cues that are shown to be helpful for discriminating IDs under the supervised setting. Various methods, which adopt direct partition~\cite{PCB,Cheng2016}, multiple granularities~\cite{MGN,zheng2019pyramidal}, attention schemes~\cite{Li2018HAN,Si2018Dual}, or transformer-based techniques~\cite{APD,Zhang2021HAT,PAT,AAformer,LA-Transformer}, have been developed to boost the performance of supervised Re-ID. In contrast, there are only a few studies on exploiting part-level features for unsupervised person Re-ID. These unsupervised methods either design part-feature based losses~\cite{Yang2019_patch,SSG,UP-ReID}, or integrate part-level features for similarity measurement~\cite{Lin2020_part}, or use part features to refine pseudo labels~\cite{PPLR}. In these unsupervised methods, part-level features are all extracted from CNN backbones. %How to extract part features from transformer networks have not been investigated yet.

%
%Part-features are obtained by directly partition of CNN backbones or a spatial transformer based network. In 

%The effectiveness of fine-grained cues has been successfully validated in supervised person Re-ID.

%The part information encodes fine-grained local cues that may be helpful in discriminate different ID and thus is frequently adopted by both supervised and unsupervised methods.
%Usually, the part information is obtained from part features which are extracted from different regions of the input image.
%Uniform partition of parts is widely used by many methods~\cite{PCB,MGN,zheng2019pyramidal,SSG,PPLR,UP-ReID}.
%As a typical part-based method, PCB~\cite{PCB} uniformly splits the feature map vertically and uses Refined Part Pooling (RPP) to re-assign outliers in each part, contributing a strong baseline model.
%Some methods consider the coarse-to-fine part features with overlapped region partition~\cite{MGN,zheng2019pyramidal}.
%In some unsupervised methods, clustering is combined with part features for better performance~\cite{SSG,PPLR} and part-level contrastive learning is also adopted~\cite{UP-ReID}.
%In addition, there are some works which do not follow the uniform partition paradigm. PL-Net~\cite{PL-Net} proposes a K-means partition method to generate part regions by the model itself.
%Auto-ReID~\cite{Auto-ReID} re-thinks from the perspective of Neural Architecture Search (NAS) with proposed part-aware module based on self-attention mechanism to search a best CNN structure.

\subsection{Transformer-based Person Re-ID}
Transformer has demonstrated its great potential in various vision tasks~\cite{ViT,SwinT}. It has also been applied to person Re-ID in recent years, mostly under the supervised setting. Initial methods~\cite{Zhang2021HAT,PAT,APD} often integrate transformer layers with CNN backbones to capture fine-grained cues and long-range contexts. For instance, TPM~\cite{APD} constructs a transformer-based module to adaptively merge parts generated from CNN-based networks such as PCB~\cite{PCB} or MGN~\cite{MGN}. HAT~\cite{Zhang2021HAT} designs a hierarchical aggregation transformer upon ResNet-50~\cite{ResNet} to integrate low-level details with high-level semantics. PAT~\cite{PAT} appends a part-aware transformer after a CNN backbone to discover diverse parts. Recently, pure transformer architectures, such as TransReID~\cite{TransReID}, AAformer~\cite{AAformer}, and LA-Transformer~\cite{LA-Transformer} are also developed for supervised Re-ID. AAformer~\cite{AAformer} introduces part tokens to learn part features in transformer while LA-Transformer~\cite{LA-Transformer} designs a PCB-like strategy to extract part-level features. In contrast to them~\cite{AAformer,LA-Transformer}, we construct a dual-branch architecture appended to a pure transformer to learn features at multiple granualities and utilize the multi-grained features for unsupervised Re-ID.

\section{The Proposed Method}
\begin{figure*}
	\begin{center}
		\includegraphics[width=0.95\textwidth]{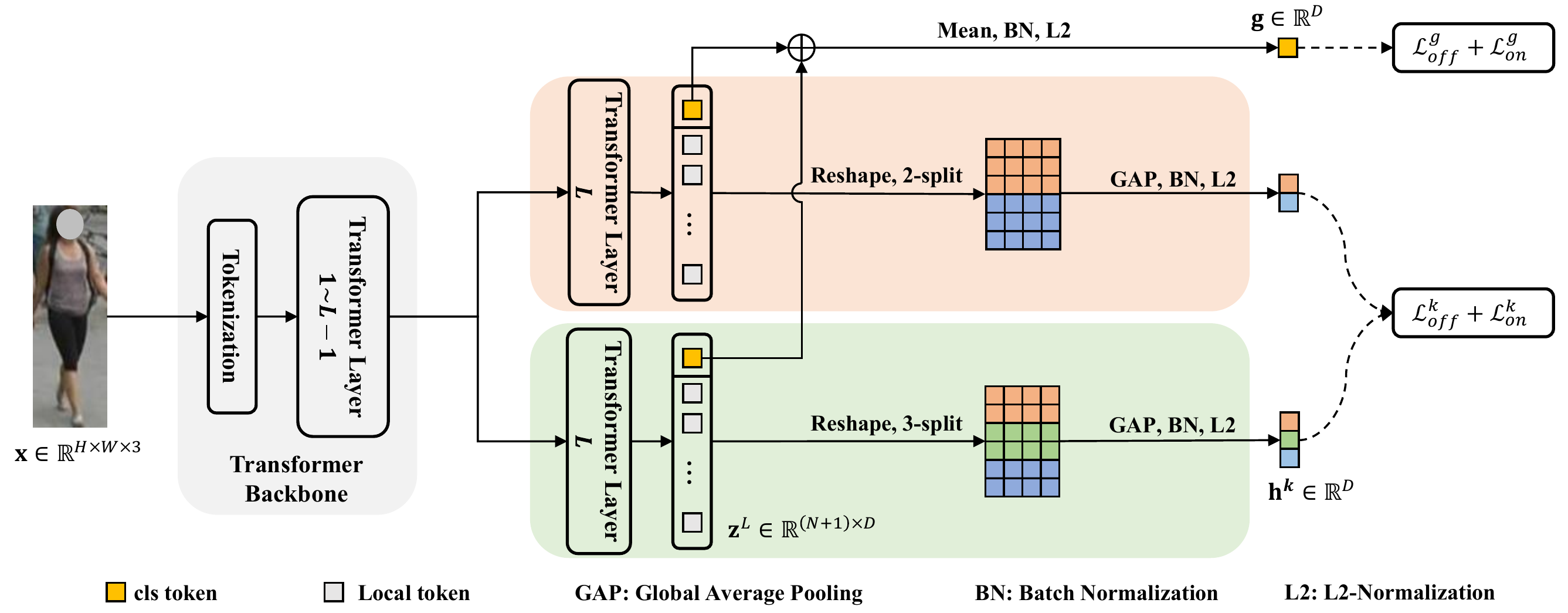}
	\end{center}
	\caption{An overview of the proposed method. It constructs a dual-branch architecture appended to a pure transformer backbone to extract features at multiple granularities. Besides, offline and online contrastive learning losses defined with respect to both global and part-level features are employed for unsupervised learning.}
	\label{fig:TMG_pipeline}
\end{figure*}

This work aims to investigate the way of extracting multi-grained features from a pure transformer network to address unsupervised Re-ID. To this end, we take a modified Vision Transformer (ViT)~\cite{ViT,TransReID,TransReID-SSL} as our backbone and construct a dual-branch architecture appended to the backbone for feature extraction. The last transformer layer is duplicated for each branch to produce outputs independently. Then, the local tokens in each branch are reshaped and uniformly partitioned into multiple stripes to yield part-level features, and meanwhile the global tokens of both branches are averaged to get a global feature. Further, based upon O2CAP~\cite{O2CAP}, offline and online contrastive learning losses with respect to both global and part-level features are defined for unsupervised learning. During test time, only the global feature obtained from two global tokens is used for inference. Figure~\ref{fig:TMG_pipeline} illustrates the entire framework.

%The entire framework is illustrated in Figure~\ref{fig:TMG_pipeline}.

%To this end, we take the TransReID-SSL~\cite{TransReID,TransReID-SSL} network as our backbone, which builds upon the Vision Transformer (ViT)~\cite{ViT} but slightly modifies ViT to adapt to the Re-ID task. Then, we construct a dual-branch architecture based on the transformer backbone. Each branch duplicates the last transformer layer to produce outputs independently. The output local tokens in each branch, which are corresponding to input patches~\cite{LA-Transformer,Beal2020}, are reshaped and uniformly partitioned into various numbers of stripes to learn part-level features at different granularities. Meanwhile, the output global tokens of both branches are averaged to generate a global feature. By this means, multi-grained features are effectively learned from the pure transformer network.
%
%To take advantage of the learned multi-grained features for unsupervised Re-ID, we choose offline-online associated camera-aware proxies (O2CAP)~\cite{O2CAP} as our learning framework. O2CAP~\cite{O2CAP} is a top-performing unsupervised Re-ID method based on a CNN backbone. We extend it by replacing its backbone with our transformer-based dual-branch network for feature extraction. In addition, we define offline and online contrastive learning losses with respect to both global and part-level features. 

\subsection{Transformer Backbone}
\begin{figure}[h]
	\begin{center}
		\includegraphics[width=0.85\columnwidth]{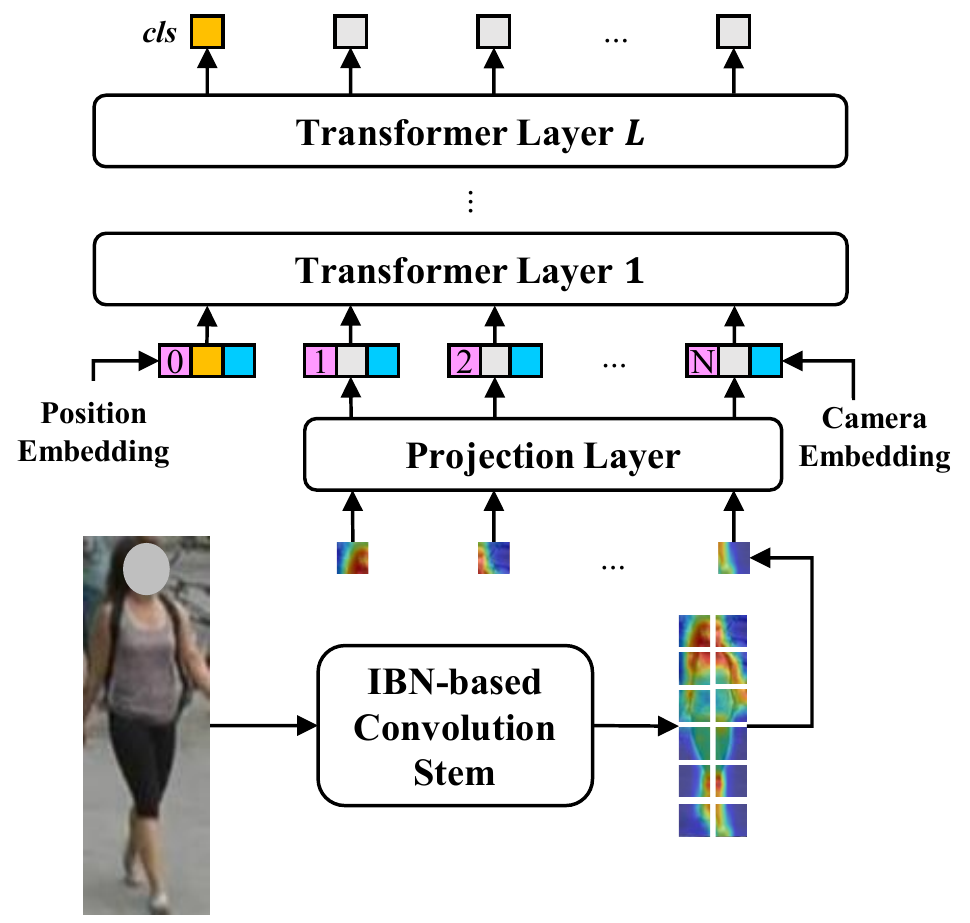}
	\end{center}
	\caption{An illustration of the transformer backbone~\cite{TransReID-SSL}. It builds upon ViT~\cite{ViT} but makes the following modifications: 1) replace the vanilla convolution stem~\cite{xiao2021early} by the IBN-based convolution stem and 2) additionally include camera embeddings.}
	\label{fig:transformer}
\end{figure}

We choose the pure transformer architecture used in TransReID-SSL~\cite{TransReID-SSL} as our backbone. It builds upon Vision Transformer (ViT)~\cite{ViT} but makes two modifications to adapt to person Re-ID. Specifically, as shown in Figure~\ref{fig:transformer}, an input image $\mathbf{x}\in  \mathbb{R}^{H \times W \times 3}$ is first passed through an Instance-Batch Normalization (IBN)-based convolution stem to produce a feature map $\mathbf{x'}\in \mathbb{R}^{\frac{H}{2} \times \frac{W}{2} \times C}$, in which $H$ and $W$ are the height and width of the image and $C$ is the number of channels. The IBN-based convolution stem, inspired by IBN-Net~\cite{Pan2018ibn} that is extensively used in CNN-based Re-ID methods, takes place of the vanilla convolution stem~\cite{xiao2021early} which is previously added in ViT to further improve training stability and generalization ability. Then, the feature map is split into $N = \frac{HW}{P^2}$ non-overlapping patches and each patch is in size of $\frac{P}{2}\times \frac{P}{2}$. Further, each patch is projected into a $D$-dimensional feature $\mathbf{f}\in \mathbb{R}^D$ as an embedded token. A learnable class token $cls$ is prepended to the sequence of patch tokens. Finally, position embeddings, together with additional camera embeddings~\cite{TransReID-SSL,TransReID}, are added to the embedded patch tokens to form the input for a transformer network. The entire procedure of tokenization is formally defined as follows:
\begin{equation}
\begin{aligned}
	\mathbf{f}_i &= \psi_i \left(\mathrm{ICS}\left(\mathbf{x}\right) \right), \ \ i = 1, ..., N; \\
	\mathbf{z}^0 &= \left[{cls};\ \mathbf{f}_1;\ ...;\ \mathbf{f}_N \right] + \mathbf{p} + \lambda_c \mathbf{c}.
	\end{aligned}
\end{equation}
Here, $\mathrm{ICS}$ denotes the IBN-based convolution stem and $\psi_i$ is the partition and projection operation. $\mathbf{p}\in \mathbb{R}^{(N+1)\times D}$ is the position embedding, $\mathbf{c}\in \mathbb{R}^{(N+1)\times D}$ is the camera embedding, and $\lambda_c$ is a hyper-parameter for weighting. 

The embedded tokens $\mathbf{z}^0$ are then input to the transformer network composed of $L$ transformer layers exactly as ViT~\cite{ViT}. Each layer consists of Multi-head Self-Attention (MSA) and Multi-Layer Perceptron (MLP) modules. The computation of layer $l \in\{1, ..., L\}$ is defined as follows:
\begin{equation} \label{eq:MSA_FFN}
	\begin{aligned}
		\hat{\mathbf{z}}^{l-1} &= \mathrm{MSA}(\mathrm{LN}(\mathbf{z}^{l-1})) + \mathbf{z}^{l-1}, \\
		\mathbf{z}^{l} &= \mathrm{MLP}(\mathrm{LN}(\hat{\mathbf{z}}^{l-1})) + \hat{\mathbf{z}}^{l-1},
	\end{aligned}
\end{equation}
in which LN denotes layer normalization. Therefore, we denote the final output of the transformer network as 
\begin{equation}
	\mathbf{z}^L = \left[{cls}^L;\ \mathbf{f}_1^L;\ ...;\ \mathbf{f}_N^L \right].
\end{equation}

%--------------------------------------
\subsection{Multiple Granularity Architecture}
\label{sec:MGA}
Inspired by the CNN-based Multiple Granularity Network (MGN)~\cite{MGN}, we construct a dual-branch architecture to extract features at multiple granularities. The architecture is illustrated in Figure~\ref{fig:TMG_pipeline}. We modify the above-introduced transformer backbone by duplicating the $L$-th transformer layer while keeping all previous layers unchanged. Each branch has one copy of the $L$-th transformer layer, which outputs one global token $cls^L$ together with $N$ local tokens $\left[\mathbf{f}_1^L;\ ...;\ \mathbf{f}_N^L \right]$. As pointed out in~\cite{LA-Transformer,Beal2020}, the local tokens substantially correspond to the original input patches and hence encode fine-grained local information. We therefore reshape the local tokens in each branch and partition them into horizontal stripes to generate part-level features. 

More specifically, we denote the output local tokens in the $i$-th branch as $\left[\mathbf{f}_{i,1}^L;\ ...;\ \mathbf{f}_{i,N}^L \right]$, which are reshaped into a feature map $\mathbf{f}'_i$ in size of $\frac{H}{P}\times \frac{W}{P} \times D$. Then, the reshaped feature map is uniformly split into $K_i$ horizontal parts and each part gets a $D$-dimensional feature by average pooling all features belonging to this part. Formally, this procedure is described as 
\begin{equation}
\begin{aligned}
\mathbf{f}'_i &= \mathrm{Reshape}\left( \left[\mathbf{f}_{i,1}^L;\ ...;\ \mathbf{f}_{i,N}^L \right] \right), \ \ i = 1, 2; \\
\mathbf{h}_{i,k} &= \mathrm{AvgPool}\left(\mathrm{Split}\left(\mathbf{f}'_i, k\right)\right), \ \ k = 1, ..., K_i.
\end{aligned}
\end{equation}
Here, $\mathbf{h}_{i,k}$ denotes the $k$-th part-level feature in the $i$-th branch. With a bit change of the subscripts, we denote the  set of all part features as $\{\mathbf{h}^k\}_{k=1}^{K}$, where $K = K_1 + K_2$.

Note that each branch also outputs a global token. Therefore, different from MGN~\cite{MGN} that additionally constructs a global branch, we get a global feature by simply averaging the global tokens produced in both branches. That is,
\begin{equation}
\mathbf{g} =  \frac{1}{2} \left(cls_1^L + cls_2^L\right).
\end{equation}
Both the global feature and part-level features are further passed through a batch normalization (BN) layer and a $L_2$ normalization layer to have unit norms.

\subsection{Unsupervised Re-ID Losses}
We intend to leverage multi-grained features to promote the performance of unsupervised Re-ID. In this work, we build our losses upon O2CAP~\cite{O2CAP}, which is a state-of-the-art unsupervised method using CNN-extracted global features. We extend it to include both global and part-level features for learning.

Given an unlabeled dataset, O2CAP performs the unsupervised learning by conducting a clustering step and a model learning step alternatively and iteratively. In each clustering step, it utilizes DBSCAN~\cite{DBSCAN} to cluster all images with respect to their global features, and then split each cluster into multiple camera-aware proxies according to camera information. The proxies are taken as pseudo labels for supervision. Let us denote the dataset with pseudo labels as $\mathcal{D} = \{(\mathbf{x}_i, \tilde{y}_i)\}_{i=1}^{N_I}$, where image $\mathbf{x}_i$ extracts a global feature $\mathbf{g}_i$ and a set of part-level features $\{\mathbf{h}_i^k\}_{k=1}^K$. Besides, $\tilde{y}_i \in \{1, \cdots, N_p\}$ is a generated pseudo label, $N_p$ is the number of proxies and $N_I$ is the number of images. Then, a proxy-level memory bank $\mathcal{K} \in \mathbb{R}^{N_p \times D}$ is constructed, in which each entry stores the centroid of a proxy's global feature. During back-propagation, when image $\mathbf{x}_i$ is input, the entry corresponding to the pseudo class $\tilde{y}_i$ is updated via
\begin{equation}
	\mathcal{K}[\tilde{y}_i] \leftarrow \mu \mathcal{K}[\tilde{y}_i] + (1 - \mu) \mathbf{g}_i,
	\label{eq:mu}
\end{equation}
where $\mathcal{K}[\tilde{y}_i]$ denotes the $\tilde{y}_i$-th entry of the memory bank and $\mu \in [0,1]$ is an updating rate. 

Assisted with the memory bank, two contrastive learning losses are designed based on offline and online associations respectively. For image $\mathbf{x}_i$, the offline association retrieves a positive proxy set $\mathcal{P}_1$ directly according to the offline clustering and splitting results while gets a negative set $\mathcal{Q}_1$ from the remaining hard negative proxies. $\mathcal{P}_1$ and $\mathcal{Q}_1$, respectively, store the indexes of associated positive and negative proxies. Then, the offline contrastive loss is defined as
\begin{equation}
\small
\mathcal{L}^g_{off}=-\sum_{i=1}^{B}\left(\frac{1}{|\mathcal{P}_1|}\sum_{u \in \mathcal{P}_1} \log \frac{S(u, \mathbf{g}_i)}{\sum\limits_{p \in \mathcal{P}_1} S(p, \mathbf{g}_i) + \sum\limits_{q \in \mathcal{Q}_1} S(q, \mathbf{g}_i)}\right),
\label{eq_inter1}
\end{equation}
in which $S(u, \mathbf{g}_i) = \exp (\mathcal{K}[u]^T \mathbf{g}_i / \tau)$, $\tau$ is a temperature factor. Moreover, $| \cdot |$ denotes the cardinality of a set and $B$ is the batch size. 

However, offline association is noisy due to the imperfect clustering results. To remedy the noise, an online association strategy is proposed. It employs an instance-proxy balanced similarity and a camera-aware nearest neighbor scheme to select a positive proxy set $\mathcal{P}_2$ and a negative set $\mathcal{Q}_2$ for each anchor image $\mathbf{x}_i$ on the fly. The online contrastive loss is defined as
\begin{equation}
\small
\mathcal{L}^g_{on}=-\sum_{i=1}^{B}\left(\frac{1}{|\mathcal{P}_2|}\sum_{u \in \mathcal{P}_2} \log \frac{S(u, \mathbf{g}_i)}{\sum\limits_{p \in \mathcal{P}_2} S(p, \mathbf{g}_i) + \sum\limits_{q \in \mathcal{Q}_2} S(q, \mathbf{g}_i)}\right).
\label{eq_inter2}
\end{equation}

%The instance-proxy balanced similarity is defined by
%\begin{equation}
%sim(\mathbf{g}_i, \mathcal{K}[j]) = w \mathbf{g}_i^T \mathcal{K}[j]  + (1-w) \mathcal{K}[\tilde{y}_i]^T \mathcal{K}[j].
%\label{eq_weighted_sim}
%\end{equation}

The above-introduced losses are defined with respect to global features. In order to leverage part-level features, we additionally construct memory banks and define both types of losses for each part, while keeping the clustering step unchanged (i.e. only using the global features for clustering as in O2CAP~\cite{O2CAP}). The entire loss for training is as follows:
\begin{equation}
\mathcal{L} = \mathcal{L}^g_{off} + \mathcal{L}^g_{on} + \lambda_p \frac{1}{K} \sum_{k=1}^{K}\left(\mathcal{L}^k_{off} + \mathcal{L}^k_{on}\right),
\end{equation}
where $\lambda_p$ is a weighting factor to balance global and part-based losses.

\section{Experiments}

\subsection{Datasets and Evaluation Metrics}
To validate the proposed method, we conduct a series of experiments on three person Re-ID datasets: Market1501~\cite{Market1501}, DukeMTMC-reID~\cite{ristani2016performance}, and MSMT17~\cite{MSMT17}. The former two datasets are captured on university campus with outdoor scenarios only, while MSMT17~\cite{MSMT17} contains both indoor and outdoor scenarios and therefore is more challenging. The number of cameras, together with the number of IDs and images contained in training, query, and gallery sets on three datasets are listed in Table~\ref{tab:datasets}. The training sets are used for unsupervised learning. During test time, each image in query is matched to similar images in gallery sets. 

%We use two datasets to train and evaluate: Market1501~\cite{Market1501} and MSMT17~\cite{MSMT17}.
%Market1501 is collected from outdoor areas of university campus, which is relatively simple. MSMT17 contains more ID categories and image samples and involves more complex illumination variations from indoor and outdoor scenarios, thus making it more challenging.
%Tab.\ref{tab:datasets} gives the basic description of the datasets.
\begin{table}[ht]
	\begin{center}
		\resizebox{\linewidth}{!}{
			\begin{tabular}{c|ccc|cc|cc}
				\hline
				\multirow{2}{*}{Dataset} &\multicolumn{3}{c|}{Training Set}           & \multicolumn{2}{c|}{Query Set} & \multicolumn{2}{c}{Gallery Set} \\ \cline{2-8} 
				& \#Camera & \#ID & \#Image & \#ID   & \#Image   & \#ID    & \#Image   \\ \hline
				Market1501               & 6              & 751        & 12,936         & 750          & 3,368            & 751           & 15,913           \\
				DukeMTMC-reID            & 8              & 702        & 16,522         & 702          & 2,228            & 1,110          & 17,661           \\
				MSMT17                   & 15             & 1,041       & 32,621         & 3,060         & 11,659           & 3,060          & 82,161           \\
				\hline
			\end{tabular}
		}
	\end{center}
	\caption{The number of cameras, IDs and images on three datasets.}
	\label{tab:datasets}
\end{table}	

As the common practice~\cite{O2CAP,ICE}, we employ the extensively used mean Average Precision (mAP) and Cumulative Matching Characteristic (CMC) for performance evaluation. The CMC metric is reported via Rank-1, Rank-5, and Rank-10. Besides, no post-processing techniques (e.g. Re-ranking~\cite{Zhong2017reranking}) are used. 

\subsection{Implementation Details}
Our transformer backbone is built upon the ViT-Small/16 model~\cite{TransReID-SSL,ViT}, which has $L=12$ transformer layers and the number of heads in each MSA is 6, the feature dimension is $D=384$. The backbone is pre-trained on a large-scale unlabeled dataset LUPerson~\cite{LUPerson}. Each image is resized to $384\times128$ and augmented with random horizontal flipping, cropping and erasing~\cite{random-erase}. Patch size $P=16$, that is, each feature map produced after the IBN-based convolution stem is split into patches in size of $8 \times 8$. The weight for camera embedding is $\lambda_c = 3$. The numbers of parts in two branches are $K_1 = 2$ and $K_2 =3$ unless otherwise specified. 

In unsupervised learning, all involved hyper-parameters are set the same as those in O2CAP~\cite{O2CAP}. That is, the updating rate $\mu = 0.2$, the temperature factor $\tau = 0.07$, and the batch size $B = 32$. Moreover, the weight of part-based losses is $\lambda_p = 0.1$. The model is trained by SGD optimizer for 50 epochs, with a momentum of 0.9, a learning rate of 0.00035 and a weight decay of 0.0005. The learning rate is regulated by a warmup scheduler that multiplies $0.01$ on the initial learning rate and linearly enlarges it to the standard value in previous 10 epochs. At epoch 20 and 40, the learning rate is divided by $10$. The entire algorithm is implemented in PyTorch~\cite{pytorch}. To accelerate training, we use half-precision floating point (FP16) computation in training while use full precision (FP32) for test. All experiments are run on a single NVIDIA GTX 1080 GPU.

\subsection{Ablation Studies}
We first conduct a series of experiments to validate the effectiveness of the proposed method, which is referred to as the Transformer-based Multi-Grained Feature (TMGF) method. Different variants of our model are investigated and all experiments are conducted on MSMT17~\cite{MSMT17}.

% designed for multi-grained feature extraction. All experiments are performed on MSMT17~\cite{MSMT17} 

% \subsubsection{Effectiveness of Two Branches}
~

\noindent\textbf{Effectiveness of Two Branches.}
In order to investigate the effectiveness of our dual-branch architecture, we compare the full model with three variants, including: 1) TMGF1: a baseline model that simply adopts the original transformer backbone and uses the global token as the image feature for learning; 2) TMGF2: a model with a single branch and the local tokens are partitioned into two parts, and meanwhile both global and part-level features are taken for learning; and 3) TMGF3: a model with a single branch similar to TMGF2 but with three partitions for local tokens. The performance of all model variants are reported in Table~\ref{tab:ablation_branch}. From the results we observe that the leverage of local tokens can consistently improve the performance, and 2-partition outperforms 3-partition if only a single branch is constructed. When two branches are working together, the mAP performance is further boosted while Rank-1 is slightly dropped (comparing TMGF vs. TMGF2).

%From the results we observe that the leverage of local tokens (e.g. TMGF2 or TMGF3 vs. TMGF) can 

\begin{table}[h]
	\begin{center}
%		\begin{tabular}{c|cc|cc}
%			\hline
%			\multirow{2}{*}{Methods} & \multicolumn{2}{c|}{Branch} & \multicolumn{2}{c}{DukeMTMC-reID} \\ \cline{2-5} 
%			&          $H_1$ & $H_2$              & mAP & Rank-1 \\
%			\hline
%			TMG w/o $H_1$+$H_2$ & & & 75.3 & 85.6 \\
%			TMG w/o $H_2$ & \checkmark &  & \textbf{77.1} & 86.8 \\
%			TMG w/o $H_1$ &  & \checkmark & 75.2 & 85.9 \\
%			TMG (full) & \checkmark & \checkmark & \textbf{77.1} & \textbf{87.1} \\
%			\hline
%		\end{tabular}
		\resizebox{\linewidth}{!}{
			\begin{tabular}{c|c|cc|cc}
				\hline
				\multirow{2}{*}{Models} & \multirow{2}{*}{Duplicate Layer L} & \multicolumn{2}{c|}{Partition} & \multicolumn{2}{c}{MSMT17} \\ \cline{3-6} 
				&    & 2-split & 3-split             & mAP & Rank-1 \\
				\hline
				TMGF1 (Baseline) &   & & & 53.5 & 81.7 \\ \hline
				TMGF2 (Single B.) &   & \checkmark & & 57.8 & \textbf{83.6} \\
				TMGF3 (Single B.) &   & & \checkmark & 54.7 & 82.0 \\ \hline
				TMGF4 (w/o Dup.) &  & \checkmark & \checkmark & 56.8 & 82.8 \\
				TMGF (Full) & \checkmark  & \checkmark & \checkmark & \textbf{58.2} & 83.3 \\
				\hline
			\end{tabular}
		}
	\end{center}
	\caption{Comparison of the proposed full model and its variants in different architectures. ``Single B.'' denotes ``single branch'' and ``w/o Dup.'' denotes ``without duplication''.}
	\label{tab:ablation_branch}
\end{table}

% \subsubsection{Effectiveness of Duplicating Layer $L$}
~

\noindent\textbf{Effectiveness of Duplicating Layer $L$.}
In our dual-branch architecture, the last transformer layer (i.e. layer $L$) is duplicated so that each branch has one copy to produce independent outputs for different partition. An alternative design is to simply append two branches after the original transformer backbone without duplication, which performs two different partitions on the same output local tokens. We refer to this design as the TMGF4 model, whose performance is also presented in Table~\ref{tab:ablation_branch}. When different partitions are conducted on the same outputs, the network may mix up the learning of local cues, leading to a degenerated performance. Therefore, it is necessary to duplicate the last transformer layer for better learning.

%It shows that duplicating the last transformer layer is necessary.

%Learning features in multiple granularities with one mixed signle branch might dilute the importance of detailed information. 

% \subsubsection{Impact of Feature Granularity}
~

\noindent\textbf{Impact of Feature Granularity.}
The granularity of part-level features may influence the performance as well. Thus, we investigate three different partition ways, which split local tokens of two branches into 2 and 3 parts, or 2 and 4 parts, or 3 and 4 parts respectively. Table~\ref{tab:ablation_granularity} lists the performance under different partitions. We observe that the best performance is achieved by the one with 2 and 3 parts in two branches. 
Splitting local tokens into more parts does not bring improvement. On the contrary, more fine-grained partition degenerates the performance. 

%The performance degenerates when local tokens are split into more parts.
%Splitting local tokens into more parts degenerates the performance.
%
%More parts do not improve the performance further. 
%
%Note that the granularity of the part feature may also influence the performance. Tab.\ref{tab:ablation_granularity} shows the performance under different granularity settings.
%As we can see, the best performance appears with $g = [2, 3]$. More parts addition does not seem to further improve the performance.
\begin{table}[h]
	\begin{center}
%		duke
%		\begin{tabular}{c|cc}
%			\hline
%			\multirow{2}{*}{Granularities} & \multicolumn{2}{c}{DukeMTMC-reID} \\ \cline{2-3} 
%			& mAP            & Rank-1           \\ \hline
%			$g = [2,3]$                     & \textbf{77.1}             & \textbf{87.1}               \\
%			$g = [3,4]$                     & 76.7               &  87.0                \\
%			$g = [2,4]$                     & 75.9               &  86.6                \\ \hline
%		\end{tabular}
%		msmt
		\begin{tabular}{c|cc}
			\hline
			\multirow{2}{*}{Granularities}  & \multicolumn{2}{c}{MSMT17}  \\ \cline{2-3} 
			 & mAP & Rank-1           \\ \hline
			$[2, 3]$       &   \textbf{58.2}           &     \textbf{83.3}        \\
			$[2, 4]$      &   57.1             &   82.9               \\
			$[3, 4]$      &   56.6             &   82.7              \\	 \hline
		\end{tabular}
	\end{center}
	\caption{The performance under different partitions. $[K_1, K_2]$ denotes that the local tokens in the first and second branches are split into $K_1$ and $K_2$ parts respectively.}
	\label{tab:ablation_granularity}
\end{table}

% \subsubsection{Impact of the Fusion of Global Tokens}
~

\noindent\textbf{Impact of the Fusion of Global Tokens.}
Each branch of our dual-branch architecture yields a global token. It implies that there are different ways to generate the global feature for learning. For example, we can choose the global token produced in either branch as the global feature, or average both global tokens as introduced in Section~\ref{sec:MGA}. In this experiment we investigate the different ways and compare their performance in Table~\ref{tab:ablation_global_feature}. The results show that the way of averaging both global tokens outperforms the other ways by a considerable margin, because it encourages a more balanced learning of two branches.

%In this experiment we investigate the way how to generate the global feature for learning. We can choose the global token output in either branch as the global feature, or average both tokens as introduced in Section~\ref{sec:MGA}. Table~\ref{tab:ablation_global_feature} presents the performance obtained in different ways. It shows that the way of averaging both global tokens outperforms the other ways by a considerable margin, because it encourages a more balanced learning of two branches.

%In the dual-branch architecture, each branch yields a global token. 
%The outputs of multi-branch projection heads contain multiple global features. Tab.\ref{tab:ablation_global_feature} shows the performance of different usage of global features from all branches. We find that the model trained with global feature from branch-2 only achieves the best mAP of 57.4\% but the full model achieves the best Rank-1 of 83.6\%.

\begin{table}[h]
	\begin{center}
%		duke
%		\begin{tabular}{c|ccc|cc}
%			\hline
%			\multirow{2}{*}{Methods}  & \multicolumn{3}{c|}{Global feature}                                               & \multicolumn{2}{c}{DukeMTMC-reID} \\ \cline{2-6} 
%			& $H_1$                     & $H_2$                     & \textit{Mean}                      & mAP            & Rank-1           \\ \hline
%			TMG w/ $\mathbf{g}^{(1)}$ & \checkmark &                           &                           & 75.8           & 86.2             \\
%			TMG w/ $\mathbf{g}^{(2)}$ &                           & \checkmark &                           & 76.5           & 87.0             \\
%			TMG (full)       &                           &                           & \checkmark & \textbf{77.1}           & \textbf{87.1}             \\ \hline
%		\end{tabular}
%		msmt
		%\resizebox{\linewidth}{!}{
			\begin{tabular}{c|c|cc}
				\hline
				\multirow{2}{*}{Models}  & \multirow{2}{*}{Global Feature}   & \multicolumn{2}{c}{MSMT17} \\ \cline{3-4} 
			                                                         & &  mAP            & Rank-1          \\ \hline
				TMGF5 &  $\mathbf{g} = cls_1^L$ &      55.9      &       82.5       \\
				TMGF6 &  $\mathbf{g} = cls_2^L$ &     56.8       &      82.8        \\
				TMGF(Full)       & $\mathbf{g} = \frac{1}{2} \left(cls_1^L + cls_2^L\right)$  &  \textbf{58.2}           & \textbf{83.3}    \\ \hline
			\end{tabular}
		%}
	\end{center}
	\caption{The performance of the models using different global features. $cls_1^L$ and $cls_2^L$ denote the global tokens output in the first and second branch respectively.}
	\label{tab:ablation_global_feature}
\end{table}

\subsection{Visualization of Attention Maps}
\begin{figure}[h]
	\begin{center}
		\includegraphics[width=0.8\columnwidth]{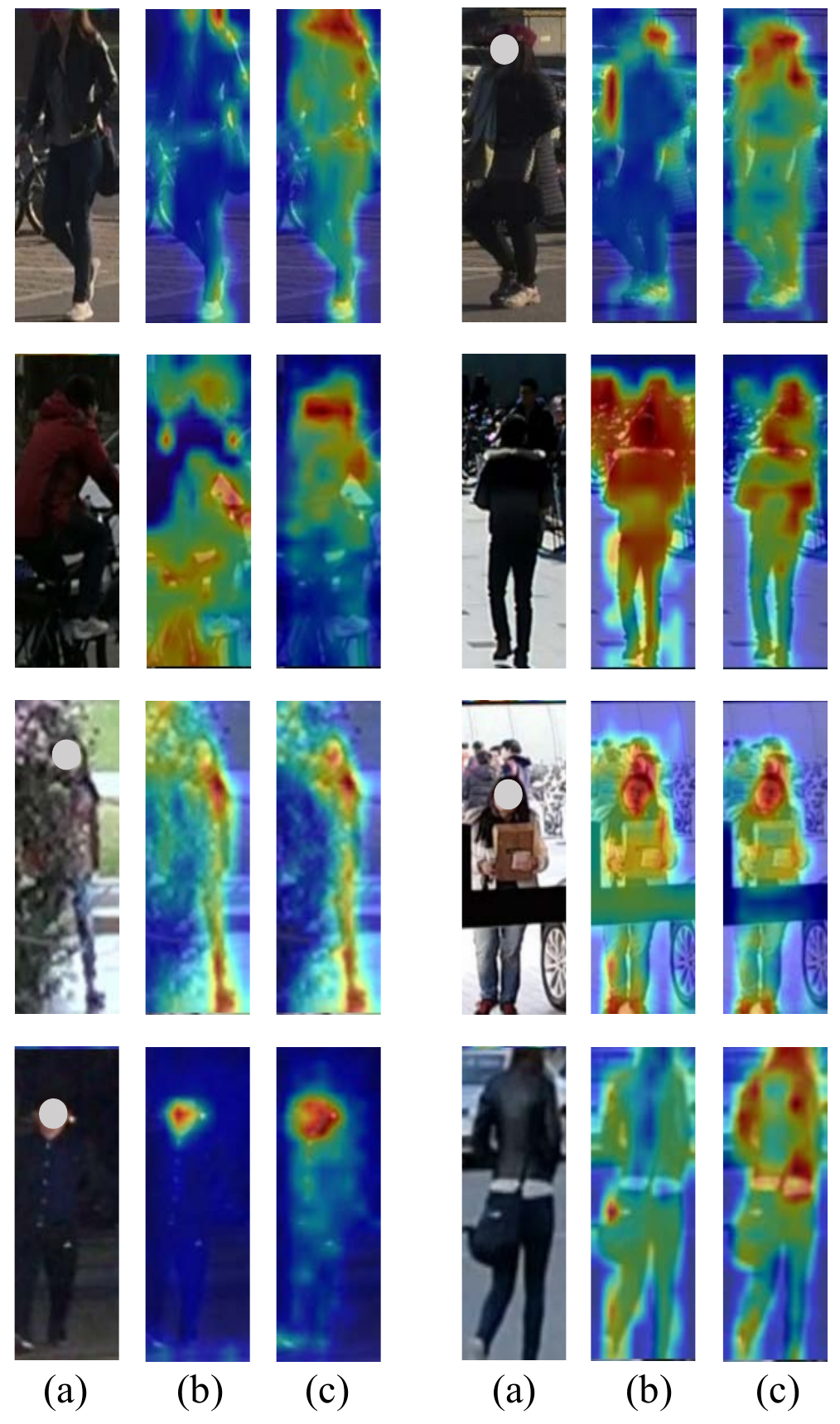}
	\end{center}
	\caption{Visualization of attention maps. (a) color images, (b) attention maps obtained by the baseline model (TMGF1), and (c) attention maps obtained by our full model (TMGF).}
	\label{fig:attention}
\end{figure}

To further illustrate the effectiveness of our dual-branch architecture, we visualize attention maps via the Attention Rollout scheme~\cite{abnar2020quantifying}. Figure~\ref{fig:attention} presents the attention maps obtained by the full model (TMGF), together with the maps obtained by the baseline model (TMGF1) for comparison. We notice that, with the exploitation of multi-grained features, our full model can attend to person regions more completely (e.g. examples in the first row) and focus more on discriminative regions (e.g. examples in the fourth row). Meanwhile, the full model is less distracted by background (e.g. examples in the second row) or occluded regions (e.g. examples in the third row). These properties enable our model to achieve a promising Re-ID performance even if the dataset is very challenging. 

%more complete regions than the baseline, while less distract by backgound or occlueded regions. 

%Figure~\ref{fig:attention} presents the attention maps obtained by the baseline model and the full model .

\subsection{Comparison to State-of-the-arts}
Finally, we compare our approach with state-of-the-art methods on three person Re-ID datasets. Table~\ref{tab:sota} presents the comparison results. 

\begin{table*}[ht]
	\begin{center}
		\resizebox{\textwidth}{!}{
			\begin{tabular}{c|c|cccc|cccc|cccc}
				\hline
				\multirow{2}{*}{Methods} & \multirow{2}{*}{Reference} & \multicolumn{4}{c|}{Market1501} & \multicolumn{4}{c|}{DukeMTMC-reID} & \multicolumn{4}{c}{MSMT17}      \\ \cline{3-14} 
				&                            & mAP & Rank-1 & Rank-5 & Rank-10 & mAP  & Rank-1  & Rank-5  & Rank-10 & mAP & Rank-1 & Rank-5 & Rank-10 \\ \hline
				% unsup
				\multicolumn{14}{l}{\textit{Purely unsupervised methods}} \\ \hline
				BUC~\cite{BUC} & AAAI19 & 38.3 & 66.2 & 79.6 & 84.5 & 27.5 & 47.4 & 62.6 & 68.4 & - & - & - & - \\
				HCT~\cite{HCT} & CVPR20 & 56.4 & 80.0 & 91.6 & 95.2 & 50.7 & 69.6 & 83.4 & 87.4 & - & - & - & - \\
				ClusterContrast~\cite{dai2021cluster} & arxiv21 & 83.0 & 92.9 & 97.2 & 98.0 & 73.6 & 85.5 & 92.2 & 94.3 & 31.2 & 61.5 & 71.8 & 76.7 \\
				CAP~\cite{CAP} & AAAI21 & 79.2 & 91.4 & 96.3 & 97.7 & 67.3 & 81.1 & 89.3 & 91.8 & 36.9 & 67.4 & 78.0 & 81.4 \\
				IICS~\cite{IICS} & CVPR21 & 72.9 & 89.5 & 95.2 & 97.0 & 64.4 & 80.0 & 89.0 & 91.6 & 26.9 & 56.4 & 68.8 & 73.4 \\
				RLCC~\cite{RLCC} & CVPR21 & 77.7 & 90.8 & 96.3 & 97.5 & 69.2 & 83.2 & 91.6 & 93.8 & 27.9 & 56.5 & 68.4 & 73.1 \\
				ICE~\cite{ICE} & ICCV21 & 82.3 & 93.8 & 97.6 & 98.4 & 69.9 & 83.3 & 91.5 & 94.1 & 38.9 & 70.2 & 80.5 & 84.4 \\
				MGH~\cite{MGH} & MM21	& 81.7 & 93.2 & 96.8 & 98.1 & 70.2 & 83.7 & 92.1 & 93.7 & 40.6 & 70.2 & 81.2 & 84.5 \\
				MCRN~\cite{MCRN} & AAAI22 & 80.8 & 92.5 & - & - & 69.9 & 83.5 & - & - & 31.2 & 63.6 & - & - \\
				MGCE-HCL~\cite{HCL} & ACPR22 & 79.6 & 92.1 & - & - & 67.5 & 82.5 & - & - & - & - & - & -  \\
				O2CAP~\cite{O2CAP} & arxiv22 & 82.7 & 92.5 & 96.9 & 98.0 & 71.2 & 83.9 & 91.3 & 93.4 & 42.4 & 72.0 & 81.9 & 85.4 \\
				PPLR~\cite{PPLR} & CVPR22 & 84.4 & 94.3 & 97.8 & 98.6 & - & - & - & - & 42.2 & 73.3 & 83.5 & 86.5 \\
				TransReID-SSL\text{\textdagger}~\cite{TransReID-SSL} & arxiv21 & \textbf{89.6} & 95.3 & - & - & - & - & - & - & 50.6 & 75.0 & - & - \\
				TMGF\text{\textdagger} & This work & 89.5 & \textbf{95.5} & \textbf{98.0} & \textbf{98.7} & \textbf{76.8} & \textbf{86.7} & \textbf{92.9} & \textbf{94.1} & \textbf{58.2} & \textbf{83.3} & \textbf{90.2} & \textbf{92.1} \\
				\hline
				% UDA			
				\multicolumn{14}{l}{\textit{UDA-based methods}} \\ \hline
				SSG~\cite{SSG} & ICCV19 & 58.3 & 80.0 & 90.0 & 92.4 & 53.4 & 73.0 & 80.6 & 83.2 & 13.3 & 32.2 & - & 51.2 \\
				SpCL~\cite{SpCL} & NIPS20 & 76.7 & 90.3 & 96.2 & 97.7 & 68.8 & 82.9 & 90.1 & 92.5 & 26.5 & 53.1 & 65.8 & 70.5 \\
				Isobe~\etal~\cite{isobe2021towards} & ICCV21 & 83.4 & 94.2 & - & - & 70.8 & 83.5 & - & - & 36.3 & 66.6 & - & - \\
				DARC~\cite{DARC} & AAAI22 & 85.1 & 94.1 & 97.6 & 98.7 & - & - & - & - & 35.2 & 64.5 & 76.2 & 80.4 \\
				\hline
				% supervised
				\multicolumn{14}{l}{\textit{Fully supervised methods}} \\ \hline
				ABD-Net~\cite{ABD-Net} & ICCV19 & 88.3 & 95.6 & - & - & 78.6 & 89.0 & - & - & 60.8 & 82.3 & 90.6 & - \\
				st-ReID~\cite{st-ReID} & AAAI19 & 86.7 & 97.2 & 99.3 & 99.5 & 82.8 & 94.0 & 97.0 & 97.8 & - & - & - & - \\
				PAT\text{\textdagger}~\cite{PAT} & CVPR21 & 88.0 & 95.4 & - & - & 78.2 & 88.8 & - & - & - &- & - & - \\
				TransReID\text{\textdagger}~\cite{TransReID} & ICCV21 & 89.5 & 95.2 & - & - & 82.6 & 90.7 & - & - & 69.4 & 86.2 & - & - \\ 
				LA-Transformer\text{\textdagger}~\cite{LA-Transformer} & arxiv21 & 94.5 & 98.3 & - & - & - & - & - & - & - & - & - & - \\
				PFD\text{\textdagger}~\cite{PFD} & AAAI22 & 89.7 & 95.5 & - & - & 83.2 & 91.2 & - & - & - & - & - & - \\
				TransReID-SSL(w/ GT)\text{\textdagger}~\cite{TransReID-SSL} & arxiv21 & 91.3 & 96.2 & - & - & - &- &- &- & 68.1 & 86.1 & - & -\\ 
				TMGF(w/ GT)\text{\textdagger} & This work & 91.9 & 96.3 & 98.9 & 99.3 & 83.1 & 92.3 & 96.4 & 97.4 & 70.3 & 88.2 & 94.1 & 95.4 \\
				\hline
			\end{tabular}
		}
	\end{center}
	\caption{Comparison with state-of-the-art methods. \text{\textdagger} indicates that the method is using a transformer backbone.}
	\label{tab:sota}
\end{table*}

% \subsubsection{Comparison with Unsupervised Methods}
~

\noindent\textbf{Comparison with Unsupervised Methods.}
We include 17 representative or recent unsupervised person Re-ID methods for comparison, among which 13 methods are purely unsupervised and 4 methods are UDA-based. Except TransReID-SSL~\cite{TransReID-SSL}, all previous methods are based on CNN backbones pre-trained on ImageNet~\cite{CNN}. It can be seen that the transformer feature extraction backbone~\cite{TransReID-SSL} greatly boosts the performance because of its network architecture and the pre-train on LUPerson~\cite{LUPerson}. Our method is built upon TransReID-SSL~\cite{TransReID-SSL}. In contrast to TransReID-SSL that uses contrastive learning losses defined in ClusterContrast~\cite{dai2021cluster}, we design a dual-branch architecture to extract multi-grained features and define both global and part-based contrastive learning losses in the form of O2CAP~\cite{O2CAP}. On the most challenging dataset MSMT17, our method surpasses TransReID-SSL by $7.6\%$ mAP and $8.3\%$ Rank-1. When compared to the CNN-based O2CAP method, we improve the performance by a significant margin. Especially on MSMT17, $15.8\%$ mAP and $11.3\%$ Rank-1 improvements are achieved.

%Compared to TransReID-SSL, our method 
%
%
 %and O2CAP~\cite{O2CAP}. 

%Except BUC~\cite{BUC}, HCT~\cite{HCT}, and SSG~\cite{SSG}, all other methods are 

%
%In this section, we compare our method with current state-of-the-art methods using both CNN and ViT backbones.
%Tab.\ref{tab:sota} shows the performances of several state-of-the-art methods. Our method achieves the performances on Market1501 and MSMT17 with mAP of 88.9\% and 57.0\%, respectively.
%Compared to O2CAP, which we take as our loss design reference, our method achieves a prominent mAP leap by 6.2\% and 14.6\% on Market1501 and MSMT17, respectively. We consider that this is due to the stronger ability of the ViT backbone to capture patch-to-patch global dependency on a image, while that in CNN backbone is restricted by the size of perception field of the convolution kernel.
%Compared to TransReID-SSL, we only get a drop on mAP by 0.7\% on Market1501 but training with less smaller batchsize of 32 (256 in TransReID-SSL). However, we achieve the state-of-the-art with great improvement of mAP by 7.1\% on the more challenging dataset MSMT17, which shows the effectiveness of the proposed method and further shrinks the gap of performance between the unsupervised and the fully-supervised methods.

% \subsubsection{Comparison with Fully Supervised Methods}
~

\noindent\textbf{Comparison with Fully Supervised Methods.}
We include seven recent fully supervised Re-ID methods for reference, in which ABD-Net~\cite{ABD-Net} and st-ReID~\cite{st-ReID} are CNN-based and all others~\cite{PAT,TransReID,LA-Transformer,PFD,TransReID-SSL} are transformer-based. Meanwhile, the performance of our model trained with ground-truth labels is also provided, which indicates the upper bound performance our method can achieve. From the results we see that our unsupervised model has already performed better than several supervised methods on Market-1501. The performance gap between our unsupervised model and the supervised ABD-Net~\cite{ABD-Net} is very small on DukeMTMC-reID and MSMT17. In addition, our model trained with ground-truth performs better than the supervised TransReID-SSL~\cite{st-ReID}, validating the effectiveness of our multi-grained feature extraction.

%LUPerson~\cite{LUPerson}

%-------------------------
\section{Conclusion}
In this work, we have presented an approach to extract multi-grained features from a pure transformer network and leverage the multi-grained features for unsupervised person Re-ID. The designed dual-branch architecture is simple but effective for feature extraction. Benefited from the multi-grained features and part-feature based learning losses, our method outperforms existing unsupervised Re-ID methods by a considerable margin, greatly mitigating the performance gap to supervised counterparts.

{\small
\bibliographystyle{ieee_fullname}
\bibliography{egbib}
}

\end{document}